\pdfoutput=1

\documentclass[11pt]{article}

\usepackage{EMNLP2022}

\usepackage{times}
\usepackage{latexsym}
\usepackage{graphicx}
\usepackage{booktabs}
\usepackage{longtable}
\usepackage{amsmath}
\usepackage{multirow}

\usepackage[utf8]{inputenc}

\usepackage{microtype}

\usepackage[T1]{fontenc}

\usepackage{inconsolata}

\title{Improving Text-to-SQL Semantic Parsing with Fine-grained Query Understanding}

\author{Jun Wang, Patrick Ng, Alexander Hanbo Li, Jiarong Jiang, \\
\textbf{Zhiguo Wang, Ramesh Nallapati, Bing Xiang, Sudipta Sengupta} \\
Amazon AWS AI Labs \\
\{juwanga, patricng, hanboli, jiarongj, zhiguow, rnallapa, bxiang, sudipta\}@amazon.com
}

\pdfoutput=1

\begin{document}
\maketitle
\begin{abstract}
Most recent research on Text-to-SQL semantic parsing relies on either parser itself or simple heuristic based approach to understand natural language query (NLQ). When synthesizing a SQL query, there is no explicit semantic information of NLQ available to the parser which leads to undesirable generalization performance. In addition, without lexical-level fine-grained query understanding, linking between query and database can only rely on fuzzy string match which leads to suboptimal performance in real applications. In view of this, in this paper we present a general-purpose, modular neural semantic parsing framework that is based on token-level fine-grained query understanding. Our framework consists of three modules: named entity recognizer (NER), neural entity linker (NEL) and neural semantic parser (NSP). By jointly modeling query and database, NER model analyzes user intents and identifies entities in the query. NEL model links typed entities to schema and cell values in database. Parser model leverages available semantic information and linking results and synthesizes tree-structured SQL queries based on dynamically generated grammar. Experiments on SQUALL, a newly released semantic parsing dataset, show that we can achieve 56.8\% execution accuracy on WikiTableQuestions (WTQ) test set, which outperforms the state-of-the-art model by 2.7\%. 
\end{abstract}

\section{Introduction}

As a natural language interface to database, Text-to-SQL semantic parsing has made great progress in recent years with availability of large amount of annotated data and advances of neural models \citep{guo2019towards, ma-etal-2020-mention, Rubin2020SmBoPSB, rat-sql, zeng-etal-2020-photon}. These models typically employ a standard encoder-decoder modeling paradigm where model first encodes query and schema, then autoregressively decodes an executable program which could be a sequence of logical form tokens for flat decoding \citep{Shi:Zhao:Boyd-Graber:Daume-III:Lee-2020} or an abstract syntax tree (AST) for structured decoding \citep{Lin2019GrammarbasedNT, rat-sql}. Either way, columns and tables are copied from input schema and literal values are copied from input query using pointer network in output program \cite{Shi:Zhao:Boyd-Graber:Daume-III:Lee-2020, rat-sql}. 

Despite the success of these models, there are several issues that are left unaddressed for real applications. 
\textbf{First}, without fine-grained query understanding, autoregressive top-down decoding suffers from generalizing to unseen query patterns at inference time \cite{herzig2020span, oren2020improving, Scholak2021:PICARD}, a problem commonly known as \textit{compositional generalization}. Specifically, model may fail to synthesize correct SQL for compound input queries like \textit{``where is the cyclist who has the most points from''} given training queries such as \textit{``where is the runner from'', ``which cyclist gets the most medals''}. 
\textbf{Second}, in previous works literal values in output logical forms are either omitted \citep{rat-sql} or directly copied from input utterances \cite{Brunner2020ValueNetAN, Shi:Zhao:Boyd-Graber:Daume-III:Lee-2020}. The former will generate non-executable queries. The latter is problematic because mentions in query are often different from their canonical forms in database. For instance, assuming a query like \textit{``how many points does LBJ get in last game''}, directly copying word \textit{``LBJ''} into SQL query won't match the name \textit{``LeBron James''} in the database.  
\textbf{Third}, as is shown in \citet{guo2019towards, Lin2019GrammarbasedNT}, structured decoding is effective in semantic parsing tasks as it's more likely to generate coherent, executable SQL queries. In structured decoding, a sequence of production rules is generated from context free grammars along with schema and literal values. However, among existing solutions, some parsers require manually designed grammars \cite{Lin2019GrammarbasedNT}. Others like RAT-SQL in \citet{rat-sql} use grammar generated from compiler tool which is often redundant, opaque to understand and offering no flexibility in model design.

To address these challenges, we propose a robust, unified framework to solve Text-to-SQL problem. Inspired by a recent work \cite{herzig2020span}, the foundation of our parser is based on fine-grained query understanding. We leverage a span based named entity recognition (NER) model to chunk input query and extract SQL-typed entities. Based on the type information we link entities to database using neural entity linker (NEL) model. NEL provides linked literal values to the parser, thus the generated SQL is executable. The final module of our framework is a grammar-based seq2seq parser which synthesizes executable logical forms from natural language query (NLQ), schema, linking results and grammar. In our parser, we dynamically build logical form grammars from training data. This approach frees us from manually constructing grammars and streamlines development of parser model. At the same time, as grammar creation is agnostic to database, our framework is more general-purpose comparing with previous works. In addition to NLQ and schema, linking information from NEL is also fed to the parser to ensure global reasoning in the decoding. Concretely this linking feature will help guide model to select proper actions in decoding. A recent work \cite{ma-etal-2020-mention} takes a similar approach as our framework --- they have an extractor model to extract entities and then link mentions to database. However, they use entity label relations to construct logic forms which greatly limits complexity of generated program. By decoupling query understanding, linking and parsing, our framework offers better explainability and flexibility in model design and optimization. 

A major challenge to build such pipeline system is fine-grained annotations which are needed to train entity recognition model and entity linking model. To tackle this issue, we leverage the newly released SQUALL dataset which provides alignment annotations between NLQ tokens and logical forms \cite{Shi:Zhao:Boyd-Graber:Daume-III:Lee-2020}. Instead of using alignment signal as attention supervision as in \citet{oren2020improving} and \citet{Shi:Zhao:Boyd-Graber:Daume-III:Lee-2020}, we programmatically convert the alignment annotations to entity annotations and linking annotations, and use these supervision signals to train NER and NEL models. A training example is shown in Figure 1 that illustrates how the aforementioned conversion works. 

\begin{figure}[t]
  \centering
  \includegraphics[width=\linewidth]{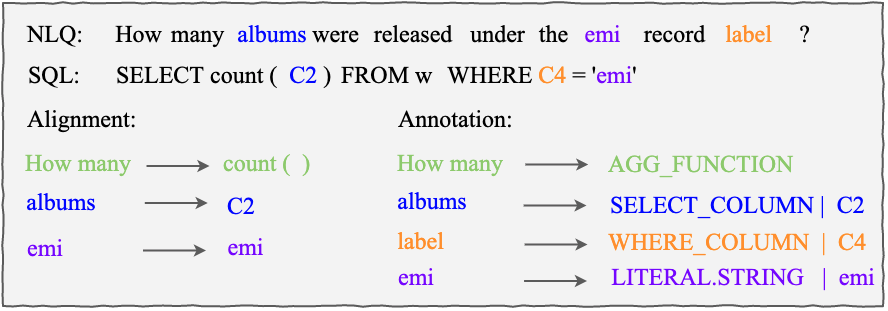}
  \caption{Fine-grained annotation from SQUALL dataset: based on target logical forms, alignment and database contents we can derive SQL-semantic annotation types for spans in query. Contents behind vertical bar in annotation are corresponding linking results.}
  \label{fig:data_format}
\end{figure}

We evaluate our framework on SQUALL dataset which has fine-grained alignment annotations. On the dev set, our framework obtains 49.36\% logical form exact match (EM) accuracy and 69.14\% execution accuracy, which is 2.2\% and 2.6\% improvement comparing with the best model in SQUALL paper, respectively. 

\section{Approach}
In this section, we describe our pipeline framework and its application to chunking, linking and parsing tasks. 

\subsection{Problem Definition}
\label{problemdef}
The input to Text-to-SQL semantic parsing problem is a sequence of natural language query tokens $Q = \{q_1, q_2, ..., q_{|Q|}\}$ and a relational database containing multiple tables $D = \{t_1, t_2, ..., t_{|T|}\}$. Each table is represented as $T = \{h_1, h_2, ... h_{|H|}, c_1, c_2, ... c_{|C|}\}$ where $h_i$ and $c_i$ are column headers and cell values in a table, respectively. The goal of the task is to generate an output program $Y$ consisting of a sequence of production rules from grammar, schema and literal values. In terms of structured decoding, an abstract syntax tree is generated and the best tree $\hat{y}$ is computed by:
\[
\hat{y} = \underset{y}{\arg\max}\mathcal\ {P( y |q, t, h, c )}
\]
given query tokens $q$ and database contents including, table names $t$, column headers $h$ and cell values $c$. Different from previous work \citep{guo2019towards}, we are targeting at generating full-fledged SQL query which is directly executable.

\subsection{Schema and Cell Value Aware NER Model}
The first stage of our framework is an NER model which serves to understand user intents in the query. Considering the fact that there could be nested entities, a span based NER model is used to chunk and identify entities in query \citep{eberts2019spanbased, zhong2020frustratingly}. We extract aggregation functions, column mentions and literal values from query. In addition, for columns we add SQL semantics to the NER tags. Specifically, as tags shown in Figure 1, we have fine-grained tags such as ``WHERE\_COLUMN", ``GROUPBY\_COLUMN" etc. A pretrained BERT base model is used as its core \citep{devlin2019bert}, as illustrated in Figure 2. 

Unlike regular NER tasks, entities in this use case highly depend on underlying database contents. To this end, we design a schema and cell value aware NER model to take database information into account. As is shown in Figure 2, we append schema and cell values to query tokens as input to the BERT encoder and separate them using ``[SEP]'' token. Let $S = \{s_1, s_2, ..., s_n\}$ denote all spans built from NLQ tokens. A span is represented as:
\[
e_s = [ e_{ctx}; e_{start}; e_{end}; e_{length} ]
\]

which is concatenation of representations of context $e_{ctx}$, start of span token $e_{start}$, end of span token $e_{end}$ and learned span length embedding $e_{length}$. The span vector then goes through a multilayer perceptron to predict whether the span is an entity and determine the corresponding entity types. We minimize the negative log-likelihood for all spans during training.
\[
p(y_s|q, t, h, c) = softmax \left( W e_s + b \right)
\]
\[
L_{NER}(\theta) =  -\sum \log p(y_s|q, t, h, c; \theta)
\]
Here $q$, $t$, $h$, $c$ have the same definition as Section \ref{problemdef} which represent query tokens, table names, table headers and cell values. $\theta$ are learnable parameters in NER model. As in \citet{zhong2020frustratingly}, a \textit{None} token is added into vocabulary of entity types. At inference time, spans which are classified as \textit{None} will be discarded. 

As the first stage of our pipeline, recall of NER model has great impact on the system performance. To improve recall performance, we introduce an additional post-processing step where we collect all schema and cell values and use them as gazetteer list. When there are \textbf{exactly matched} spans in NLQ, we force model to generate a valid entity type for such spans. At the same time, if a span overlaps with a gazetteer matched span, we choose to keep gazetteer matched span as it is more likely to be a valid span.

To further leverage matching information, we add constrained decoding after filtering \cite{lester-etal-2020-constrained}. Concretely in decoding process we force model to predict labels based on gazetteer matching category. For instance, when chunking query \textit{``how many points does LeBron James get in last game''}, if there is an exact match of span \textit{``LeBron James''} with an entry in cell value gazetteer list, the decoding logic will force the model to give a prediction tag which is compatible with cell value type. 

\begin{figure*}[t]
  \centering
  \includegraphics[width=0.85\linewidth]{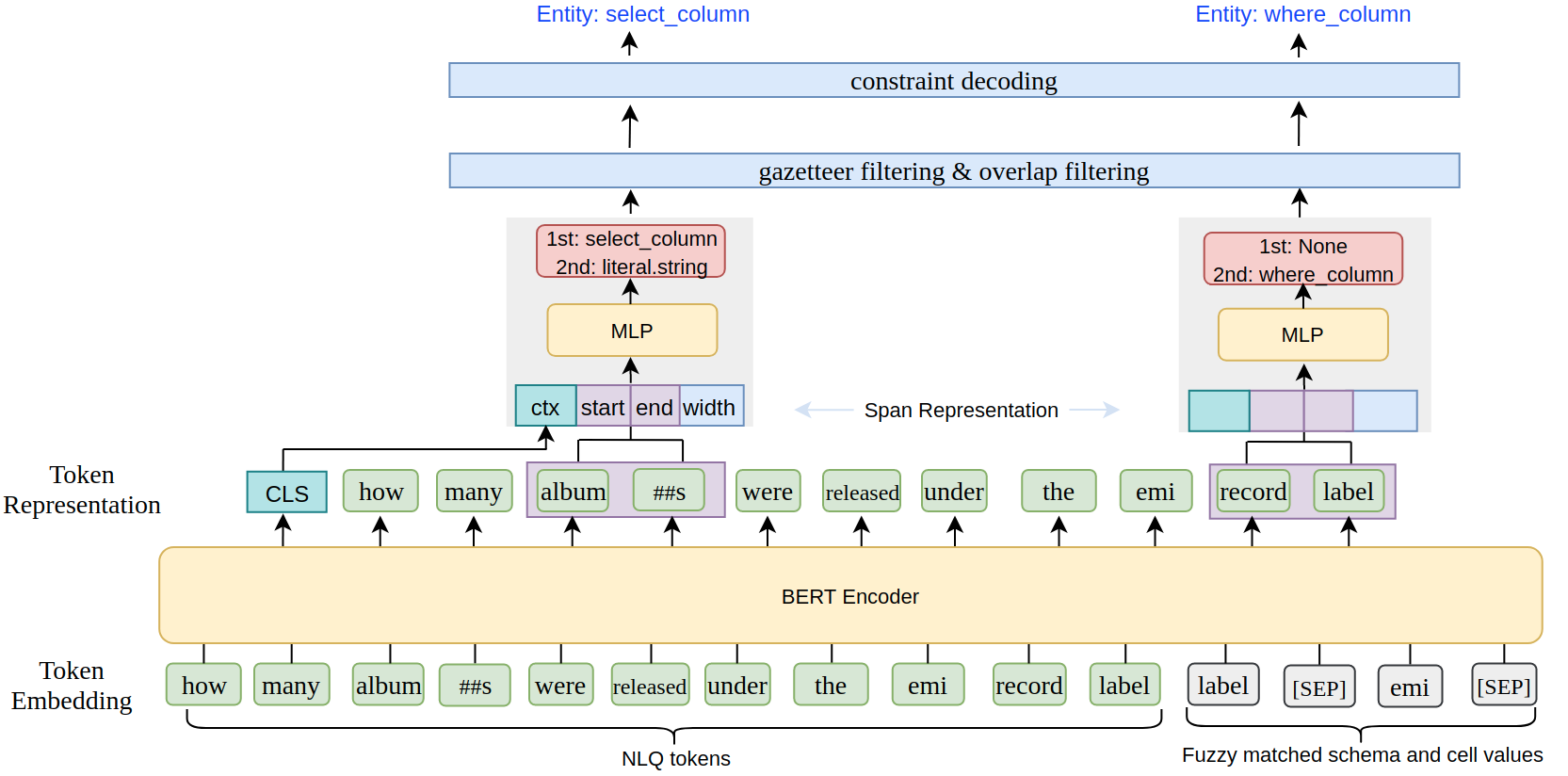}
  \caption{Span based schema and cell value aware NER model architecture. Input tokens are tokenized by BERT tokenizer.}
  \label{fig:speech_production}
\end{figure*}

\subsection{Neural Entity Linking (NEL) Model}
\label{nel}
The sketch of generated SQL logical forms consists of rules from grammar. To populate the sketch we use a pointer network in parser to copy table names, column names and cell values from input schema and query. However, directly copying these entities will lead to non-executable SQL as columns and literal values in NLQ can be different from their canonical form in the database. Our entity linking model bridges the gap between entity mentions in query and entity values in SQL. Even though fuzzy string match is widely used for linking task in literature \citep{rat-sql, Shi:Zhao:Boyd-Graber:Daume-III:Lee-2020}, in real application purely relying on fuzzy match could lead to suboptimal performance. For instance, in SQUALL dataset, there are less than 50\% of entities that have exact matches in database. In light of this, we use a neural ranking model for entity linking task. Specifically, given a mention in NLQ and a list of candidates in database, NEL model selects the best matched candidate for each entity \citep{wu2019zero}. 

For column and literal value entities from output of NER model, we construct an input to the NEL model using NLQ tokens, mention and candidates. For example, input to NEL model can take the following format:

\begin{center}
{\small \textit{NLQ} [SEP] \textit{Mention} [SEP] \textit{Candidate}} \openup .2em
\end{center}

As we know the mention type from NER model, we could narrow down candidates to a specific type. For instance, if a mention is literal value type, then candidates are only limited to cell values. It is worth to note that without fine-grained query understanding, for each mention in NLQ, linking candidates have to be all contents in database. Thus it is challenging for end-to-end model to deal with cases where there are overlapping column names and cell values. In addition, we could append additional meta features to the input of NEL model. When a mention is column type, we could construct the following input:

\begin{center}
{\small \textit{NLQ} [SEP] \textit{Mention} [SEP] \textit{Candidate} [SEP]} \openup .2em \textit{value} [SEP] \textit{column type}
\end{center}
where value is the cell value in current candidate column which has the highest fuzzy matching score with NLQ. Column type could be meta information such as data type \textit{integer}, \textit{string} etc. 

In our experiments, we use BERT base model for linking model. After BERT encoding, a linear layer is applied on the classification token ``[CLS]'' to produce a logit score for each candidate. During training, cross-entropy loss is calculated over all the linking candidates. During inference, each linking candidate is fed to the BERT model independently and is scored by BERT model. Then best candidate is selected based on ranking scores.

\subsection{Neural Semantic Parsing (NSP) Model}
\label{nsp}
Neural semantic parser model takes as input NLQ, a database and outputs a sequence of production rules which can be used to deterministically build up an output program. The backbone of our neural semantic parser is a BART Large model which is pretrained with large amount of parallel text data for both its encoder and decoder \citep{lewis-etal-2020-bart}.

\paragraph{\textbf{Encoder}} Encoder encodes NLQ tokens and schema information. Specifically, NLQ tokens, table names and column names are concatenated together with a ``[SEP]'' token used as separator. As we have already got literal value spans from NER stage, at the output of BART encoder, we collect representations for all of these literal spans by pooling average token representations in the span. 

From query understanding, we have SQL-semantic tags for each column mentions. For example, in Figure 1 we know that ``album'' is a ``SELECT\_COLUMN''. At the same time, in NEL results we know ``album'' is linked to column ``C2''. Consequently, we know that ``C2'' is used in the logical forms as a selection column. In order to utilize this information in parser, we have a column type embedding layer in the encoder. When constructing column representations, we concatenate column type embeddings to the original column representations. To alleviate upstream errors, during parser model training, we randomly drop column type feature for 20\% of time so that when NER model gives incorrect predictions, parser model learns to handle these cases. In our experiments, we will show that this feature can give big boost to parser performance.

\paragraph{\textbf{Decoder}}
The generated program at the output of a semantic parser can be a sequence of logical form tokens or an AST tree. The former decoding is generally referred to as flat decoding and the latter one is called structured decoding. To synthesize syntactically correct SQL program, in our framework a grammar based autoregressive top-down decoder is utilized to generate AST. Contrary to implementation in \citet{yin-neubig-2017-syntactic} where AST grammar is collected through compiler's tool, we dynamically generate context free grammar from training data. Concretely, during training stage, we collect ground truth SQL queries and parse them into SQL trees. Then we collect all the rules as our grammar using breadth-first search algorithm. The distinguishing feature of our grammar generation comparing with the one discussed in \citet{Lin2019GrammarbasedNT} is that we don't need so-called \textit{"linked rules"} because we get linked entity from NEL results. This method saves us from manually writing rules for each dataset. At the same time, it decouples SQL grammar from domain knowledge which makes our framework more general-purpose than previous works.

At each decoding step, decoder takes as input previous decoding result and iteratively apply production rules to non-terminal nodes. Owing to our query understanding based framework, we are able to employ soft copy mechanism \citep{See2017GetTT} to directly copy tables, columns and values from output representations of encoder. Finally, beam search is used during inference time.

\section{Experiment}

\subsection{Data}
SQUALL \citep{Shi:Zhao:Boyd-Graber:Daume-III:Lee-2020} is collected based on WikiTableQuestions which is a question-answering dataset over structured tables. In SQUALL, each query only relies on one table to get the answer. 

To obtain supervision labels for NER and NEL model, we first parse ground truth SQL queries into trees. Then based on alignment information provided by the dataset, we programmatically derive entity labels and linking labels for each entity span in the query. In total, there are 11276 training instances and 4344 testing instances in the dataset. Train, dev, test set partition is based on the pre-defined setting in the released dataset. We use logical form exact match accuracy and execution accuracy as our evaluation metric.

\subsection{Model Analysis}
We first explore the best setup for our framework. In these experiments, we always use structured decoding in the parser. Based on how to utilize query understanding results, there are three different configurations of our parser model: (1) In our baseline model setup, we only utilize NER and NEL results for entity linking purpose and copy linked results into the generated AST. (2) Instead of using all schema in parser's encoder, we only input linked columns to the parser (3) We inject fine-grained query understanding results in the parser, i.e. add linked column type information as meta features to the encoder of NSP. 

In Table~\ref{tab:system_conf}, we summarize performance of our system on dev set under different parser configurations. Comparing ``(2) Linked columns only'' model with baseline model (1), we can see that system performance suffers because when we only use linked columns in the parser, NER model errors will propagate to the downstream. With adding column type feature based approach (row 3 in Table~\ref{tab:system_conf}), we find that it can greatly boost model performance as it guides parser to choose correct columns. We also have an oracle experiment where we use ground truth column type feature. As is shown in last row in Table~\ref{tab:system_conf}, it's around 19\% better than our best configuration (row 3 in Table~\ref{tab:system_conf}) which means there is still room to improve our system. 

\begin{table}
{\tablesize
\centering
\begin{tabular}{lccc}
\toprule
\multirow{2}{*}{Model} & \multicolumn{2}{c}{Dev} \\ 
                       & $\mathrm{ACC_{LF}}$  & $\mathrm{ACC_{EXE}}$   \\ 
\midrule
(1) Baseline  & 42.56 & 60.57  \\
(2) Linked columns only  & 38.69 & 56.83  \\
(3) Columns type feature & 49.38 & 69.14  \\
(4) Oracle column type feature & 68.21 & 85.16 \\
\bottomrule
\end{tabular}
\caption{System performance with different approaches to utilize query understanding results in semantic parser. $\mathrm{ACC_{LF}}$, $\mathrm{ACC_{EXE}}$ are logical form accuracy and execution accuracy, respectively.}
\label{tab:system_conf}
}
\end{table}

In Table~\ref{tab:performance_comp}, we compare our model performance with the best model (ALIGN) in SQUALL paper. Their end-to-end model leverages BERT as the encoder and LSTM as decoder using flat decoding strategy. They're using supervised attention to help model learn alignment information. In order to have a fair comparison with SQUALL paper, we add two more baselines in our experiments: in the first experiment we augment ALIGN model by replacing BERT encoder and LSTM decoder with BART model; in the second experiment, we replace structured decoding with flat decoding in our system. In a nutshell, we compared performances of four models: (1) original ALIGN model from \citet{Shi:Zhao:Boyd-Graber:Daume-III:Lee-2020}, (2) our augmented implementation of ALIGN, (3) our framework with flat decoding, (4) our best configuration (row 3 in Table~\ref{tab:system_conf}). As can be seen from the table, our best model outperforms ALIGN model and augmented ALIGN model in both logical form accuracy and execution accuracy. On test set we achieve 56.8\% execution accuracy which is 2.7\% higher than the ALIGN model. If we compare two ALIGN models and two of our systems with different decoding strategy, it's easy to tell BART model and structured decoding contribute limited benefits in our framework which in turn suggests that major improvement in our system is from fine-grained query understanding part.
\begin{table}
{\tablesize
\resizebox{\columnwidth}{!}{%
\centering
\begin{tabular}{lccc}
\toprule
\multirow{2}{*}{Model} & \multicolumn{2}{c}{Dev} & Test\\ 
                       & $\mathrm{ACC_{LF}}$  & $\mathrm{ACC_{EXE}}$  & $\mathrm{ACC_{EXE}}$ \\ 
\midrule
(1) ALIGN(SQUALL)  & 47.2 & 66.5 & 54.1 \\
(2) ALIGN(SQUALL) + BART  & 47.7 & 67.1 & 54.6 \\
(3) Ours + Flat decoding  & 49.1 & 68.8  & 56.2 \\
(4) \textbf{Ours} & \textbf{49.4} & \textbf{69.1} & \textbf{56.8} \\
\bottomrule
\end{tabular}
}
\caption{Performance comparison of our model with ALIGN model in SQUALL paper. ALIGN + BART model is our implementation of ALIGN model where we replace BERT encoder and LSTM decoder with BART encoder and decoder.}
\label{tab:performance_comp}
}
\end{table}

We also evaluate the generalizability of our model. Our assumption is that span based NER model can chunk query based on how meaning is composed. As we add NER label information into parser, we hypothesize that the parser can learn to generate program based on semantic type of query tokens rather than just using lexical meaning of tokens. Thus, we would see better performance on compositional generalization. Particularly, we want to see how model works on nested queries since it is an ideal set to evaluate model's ability on generalization. To this end, we collect all nested queries in dev set and evaluate our system on this nested query set. The performance is shown in Table~\ref{tab:composition}. Our model shows 4.5\% improvement comparing with original ALIGN model and around 3\% improvement comparing with augmented ALIGN model. The results demonstrates that our framework is better at compositional queries.
\begin{table}
\centering
{\tablesize
\begin{tabular}{lrl}
\toprule
Model & $\mathrm{ACC_{LF}}$  \\ 
\midrule
ALIGN(SQUALL) & 30.29  \\
ALIGN(SQUALL) + BART & 30.94 \\
Ours & \textbf{34.78} \\
\bottomrule
\end{tabular}
\caption{Model's performance on nested queries. We retrained ALIGN model for this experiment.}
\label{tab:composition}
}
\end{table}

\begin{table}[!htbp]
{\tablesize
\centering
\begin{tabular}{lcc}
\toprule
\multirow{2}{*}{Model} &  \multicolumn{2}{c}{Dev Set}\\
                       &  NER F1 & $\mathrm{System \ ACC_{LF}}$ \\
\midrule
Our best model  &  \textbf{85.14}  &  \textbf{49.38}   \\
-cell   &  84.23 & 47.51   \\
-schema  &  83.70 & 47.66   \\
-gazetteer   &  84.15  &  45.89   \\
-gazetteer-cell-schema  &  82.98 & 43.66   \\
\bottomrule
\end{tabular}

\caption{NER performance ablation study. Baseline model here is span-based schema and cell value aware NER model. We gradually remove each component to see its impact on the system performance. }

\label{tab:nertable}
}
\end{table}

\subsection{Ablation Study}
As the first module of our framework, NER model plays a critical role in our pipeline system. To qualitatively study the impact of NER model, we did ablation experiments to study how each component in NER model affects the system performance. Concretely, we use our schema aware and cell value aware NER model as the baseline and gradually remove each component to see how system performance fluctuates. Table~\ref{tab:nertable} summarizes our findings. As is shown in the table, when we remove cell values, schema and gazetteer filtering, NER F1 score goes down and system performance degrades accordingly. We can also see that among these three components, system performance drops the most (from 49.38 to 45.89) when we remove gazetteer filtering. Gazetteer filtering in NER serves the role to combine string match and model prediction. It forces NER model to output predictions for exactly matched spans at output which increases recall of NER model. From system perspective, downstream parser is more sensitive to missing entities. Thus, improving NER recall can greatly boost system performance.

\section{Conclusion}
In this work, we proposed a novel, general-purpose Text-to-SQL semantic parsing framework which is based on fine-grained query understanding. The framework tackles several pain points in the Text-to-SQL problem and offers a new robust approach for real-life applications. Our framework outperforms previous state-of-the-art result by 2.7\% on SQUALL test set. In the future, we plan to explore using fine-grained query understanding results to constrain decoding search space in parser to further improve system performance. 

\section{Limitations}
While this work aims to improve Text-to-SQL semantic parsing with fine-grained annotations, we don't have enough time and resources to collect a dataset for such purpose. Due to this issue, our experiments are limited to SQUALL dataset. In the future, we plan to build a comprehensive dataset to facilitate research in the area.

\bibliography{emnlp22,custom}
\bibliographystyle{acl_natbib}

\end{document}